\documentclass[conference]{IEEEtran}
\IEEEoverridecommandlockouts
\usepackage{cite}
\usepackage{amsmath,amssymb,amsfonts}
\usepackage{algorithmic}
\usepackage{graphicx}
\usepackage{hyperref}
\usepackage{textcomp}
\usepackage{xcolor}
\usepackage[square, numbers, comma, sort&compress]{natbib}
\def\BibTeX{{\rm B\kern-.05em{\sc i\kern-.025em b}\kern-.08em
    T\kern-.1667em\lower.7ex\hbox{E}\kern-.125emX}}
\begin{document}

\title{Detecting Insincere Questions from Text: A Transfer Learning Approach}

\author{\IEEEauthorblockN{ Ashwin Rachha}
\IEEEauthorblockA{\textit{Dept. of Computer Science} \\
\textit{Pune Institute of Computer Technology,}\\
Maharashtra,India  \\}
 \href{mailto:ashwin.rachha@gmail.co}{ashwin.rachha@gmail.com}
\and
\IEEEauthorblockN{ Gaurav Vanmane }
\IEEEauthorblockA{\textit{Dept. of Computer Science} \\
\textit{Pune Institute of Computer Technology,}\\
Maharashtra,India  \\}
 \href{mailto:guaravvanmane94@gmail.com}{guaravvanmane94@gmail.com}
}

\maketitle

\begin{abstract}
 The internet today has become an unrivalled source of information where people converse on content based websites such as Quora, Reddit, StackOverflow and Twitter asking doubts and sharing knowledge with the world. A major arising problem with such websites is the proliferation of toxic comments or instances of insincerity wherein the users instead of maintaining a sincere motive indulge in spreading toxic and divisive content. The straightforward course of action in confronting this situation is detecting such content beforehand and preventing it from subsisting online. In recent times Transfer Learning in Natural Language Processing has seen an unprecedented growth. Today with the existence of transformers and various state of the art innovations, a tremendous growth has been made in various NLP domains. The introduction of BERT has caused quite a stir in the NLP community. As mentioned, when published, BERT dominated performance benchmarks and thereby inspired many other authors to experiment with it and publish similar models. This led to the development of a whole BERT-family, each member being specialized on a different task. In this paper we solve the Insincere Questions Classification problem by fine tuning four cutting age models viz BERT, RoBERTa, DistilBERT and ALBERT 
\end{abstract}


\section{Introduction}
Content based websites such as Quora, Reddit, StackOverflow are primarily used for seeking genuine answers to questions. People from different domains put up their questions and educators or people knowledgeable in a certain field answer them. One major impediment to a plain sailing execution of information exchange is the proliferation of toxic comments. The key challenge is to weed out such toxic comments termed as Insincere Questions. An Insincere Question is designated as a comment intended to make a statement than to look for genuine answers.

An Insincere Question is characterised by:
\begin{itemize}
    \item Having a non-neutral tone.
    \item Being disparaging or inflammatory.
    \item Not being grounded in reality. 
    \item Expressing  explicit content. 
\end{itemize}

This major class of problem pertains to Text classification which has been a benchmark problem of evaluating various research advancements in natural language processing. While traditional machine learning algorithms such as naive bayes, logistic regression and decision trees can be rightfully applied to this problem, they suffer with major impediments in their constructs. Vanilla RNNs, Gated Recurrent Unit and Long Short Term Memory Networks replaced their usage as the new state of the art. Even though LSTMs and GRUs performed well, they failed to capture the dependencies in long range sentences.
Now with the advent of Transfer Learning, Language model pre-training has proven to be useful in learning universal language representations. Researchers in the field are developing new and better language models at an unprecedented speed. Applying these new state of the art models could improve current methods and replace manual labeling tasks for text classification, but also find widespread application in similar other fields, such as machine translation and question answering. In this paper, we test this by applying new transformer models from the BERT-family to improve the current method of binary text classification in the context of Insincere Questions Classification. We make use of the Quora Insincere Questions Classification dataset \cite{Quora} for this purpose We find that all of our models achieve remarkable results in classifying the given  data (AUC’s ranging from 0.92-0.97), with BERT achieving the best results compared to RoBERTa, DistilBERT, and ALBERT. This indicates that the models are well equipped to take over tasks that researchers have previously solved in less optimal ways.

\section{Related Work}
Detecting divisive and inappropriate content is a highly relevant task in Natural Language Processing today. Over the last few years, the class of problems pertaining to text classification have been dominated by deep learning based architectures. Kim \cite{kim2014convolutional}, reports a series of experiments with CNN trained on top of pre trained word-vectors for sentence level classification tasks. Liu et al \cite{liu2017deep} develop what is called the XMLCNN built on top of KimCNN with modifications such as wider convolutional kernels, adaptive dynamic max-pooling, and an additional bottleneck layer to capture the features of large documents more efficiently. 
Yang et al \cite{yang2016hierarchical} propose the Hierarchical Attention Network model which consists of two levels of attention mechanism operating at the word as well as the sentence level thus focusing distinctively on more and less important content for deriving document representations.
Adhikari et al \cite{adhikari2019rethinking} propose a simple Bidirectional LSTM with attention mechanism and appropriate regularization techniques to yield next to state-of-the-art results on document classification.

Alternatively, substantial work has been done to prove pre-trained models trained on a large corpus of data yield promising results in this domain, thus  avoiding training models from scratch. Mikolov et al \cite{mikolov2013distributed} introduced a skip gram model with negative sampling termed word2vec and pennington et al \cite{pennington2014glove} came up with the glove embeddings which leverages statistical information by training only on the nonzero elements in a word-word co occurrence matrix rather than their sparse representations. Peters et al \cite{peters2018deep} introduce Elmo which is a deep contextualized word representation which models complex syntax and semantics, trained with Bidirectional LSTMs on a large text corpus using a fixed embedding for each word,  looks at the entire sentence before assigning each word in it an embedding.

More recently the attention algorithm was introduced that completely changed the landscape in the NLP field. It was first introduced by the Google Brain team with the paper “Attention is all you need” (Vaswani et al \cite{vaswani}) emphasizing the fact that their model does not use recurrent neural networks at all.  Attention had already been a known idea used in LSTMs, but this was the first time it completely took the place of the recurrence. In the following paragraphs we explore various models and advancements that are built on top of the aforementioned works.

\section{Methodology}

\subsection{Dataset}
\begin{figure}[h]
    \centering
    \includegraphics[width=8cm, height=4.5cm]{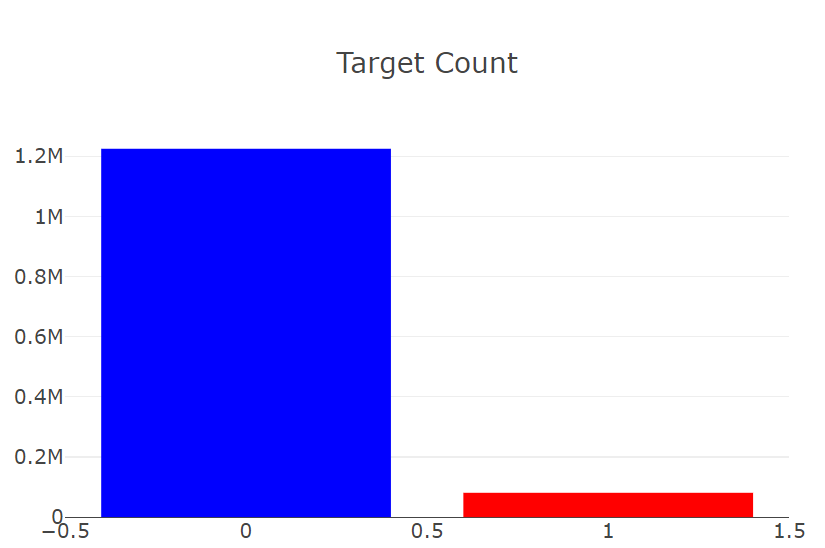}
    \caption{Quora Insincere Questions Dataset}
    \label{dataset}
\end{figure}

The train set comprises over 1.3M question texts along with their corresponding labels. A question may be sincere or insincere. Sincere questions are labeled 0 while insincere questions are labeled as 1. There is a severe class imbalance as can be seen in Fig\ref{dataset}. Values pertaining to sincere questions comprise of a large amount while only a small fraction that is 80K pertaining to insincere questions. The question texts and labels are extracted out from the train set and pre-processed to further train the models upon. 

\subsection{Data Preprocessing}
As in all language models, text data has to be preprocessed before it gets fed into the model. Bert and bert based models require its inputs to be tokenized with the BertWordPiece tokenizer which tokenizes the input sequences by matching the words with its inbuilt vocabulary and segments an individual word till it is not found in the vocabulary and completes the segmented part with \#\#. Furthermore the sequence length must be the same for each question. The max sequence length permissible for Bert is 512 but for the sake of maximum memory usage we use the length of 192. Sequence lengths greater than 192 are truncated and lesser are padded with zeros. After chunking the sequences they are encoded by the tokenizer and ready to be used. 

\subsection{Model Training}
Four transformer models viz. Bert, Roberta, DistilBert and Albert have been fine tuned on the aforementioned dataset. Training was done using the python programming language on top of the Tensorflow library and the keras framework. The training of the models was done on Tensor Processing Units (TPUs) provided by Kaggle. TPUs provide a better performance with respect to Tensorflow and Keras computations on tensors against GPUs. For efficient usage of memory the batch size was kept constant as 16 and the max length of an input sequence was initialized to 192. This made sure that the memory was completely utilized. For optimizing parameters of the model we used the Adam optimization function with a learning rate of ``1e-5 '' for all models. Since this problem relates to binary classification we used the binary cross entropy loss function. The models were trained for 3 epochs. Because of severe imbalance in the data, along with traditional validation metrics, we included AUC scores and F1 scores for validation of the models.

\section{Background}
In this section we will see various models used for traditional natural language processing and how the existing state of the art models are built on top of them.
\subsection{Recurrent Neural Networks:}
Recurrent Neural Networks are a class of artificial neural networks that were first introduced to manipulate sequential data. RNNs exhibit temporal dynamic behavior which allow them to transfer information more easily from one time step to another. The information from the previous states is termed as a hidden state.

\begin{figure}[h]
    \centering
    \includegraphics[width=8.5cm, height=3cm]{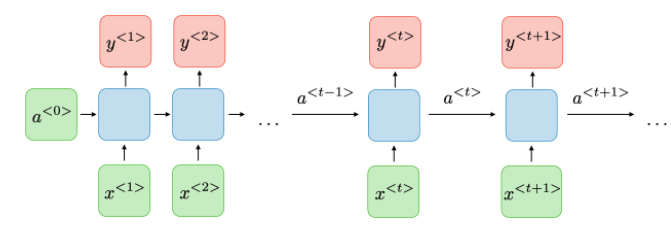}
    \caption{Architecture Of Traditional RNN}\cite{stanford.edu}
    \label{RNN}
\end{figure}

For each timestep t, the activation 
\[a^{<t>}=g_{1}\left(W_{a a} a^{<t-1>}+W_{a x} x^{<t>}+b_{a}\right)\] And output \[y^{<t>}=g_{2}\left(W_{y a} a^{<t>}+b_{y}\right)\]
where $W_{a x}$, $W_{a a}$, $W_{y a}$, $b_{a}$, $b_{y}$ are coefficients that are shared temporally and$g_{1}$,$g_{2}$ activation functions.

RNNs showed good results in a majority of applications ranging from time series prediction, text generation, biological modeling, speech recognition etc. However vanilla RNNs suffer from a major problem of vanishing gradients. As many other machine learning algorithms, RNNs are optimized using Back Propagation and due to their sequential nature the error decays severely as it propagates back through layers. The gradient which is very small, thus effectively prevents the weight from changing its value. In the worst case the neural network may even stop training further. 
\subsection{LSTM}
In order to remedy the vanishing gradients problem, Long Short Term Memory Networks were introduced, the architecture of which consists of different gates viz update, relevance, forget and output. This introduction of a tweaking in the architecture of RNNs enabled LSTMs to remember relevant context from long range dependencies and introduced the flexibility to use long sentences.
\begin{figure}[h]
    \centering
    \includegraphics[width=7cm, height=5cm]{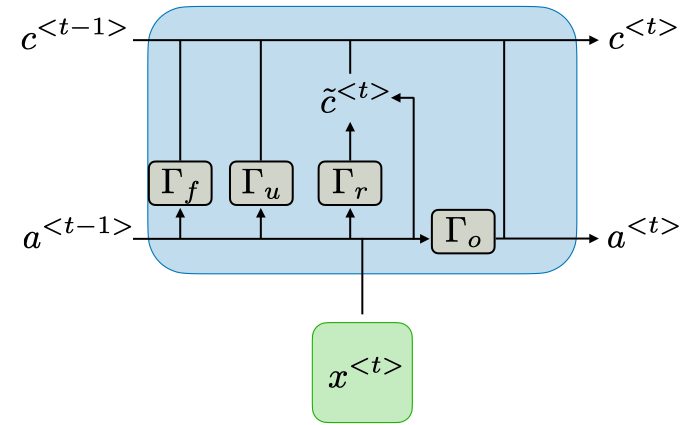}
    \caption{Long Short-Term Memory(LSTM)}\cite{stanford}
    \label{LSTM}
\end{figure}
\[\tilde{c}^{<t>}=\tanh \left(W_{c}\left[\Gamma_{r} \star a^{<t-1>}, x^{<t>}\right]+b_{c}\right)\]
\[c^{<t>}=\Gamma_{u} \star \tilde{c}^{<t>}+\Gamma_{f} \star c^{<t-1>}\]
\[a^{<t>}=\Gamma_{o} \star c^{<t>}\]
where

\[\Gamma_{u}=Update gate\]
\[\Gamma_{r}=relevance gate\]
\[\Gamma_{f}=Forget gate\]
\[\Gamma_{0}=Output gate\]

\begin{figure*}
    \centering
    \includegraphics[width=13cm, height= 12cm]{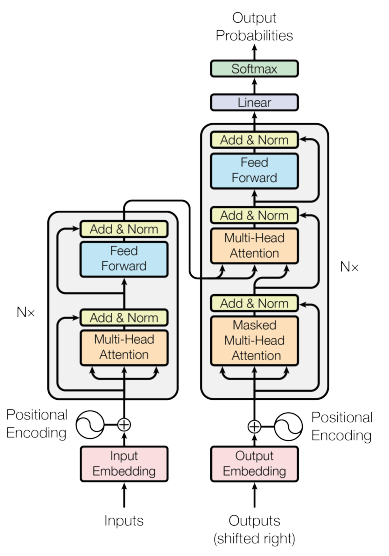}
    \caption{Transformer Architecture}\cite{vaswani}
    \label{transformer}
\end{figure*}
To boost the performance of existing architectures Bidirectional LSTMs were introduced that can process sequences from forward as well as backward directions intended for the network to learn better understandings from both sequence directions. The downside to this is the model being extremely slow due to the huge number of parameters to train. This is mitigated by the use of Gated Recurrent Networks (GRUs) as they are a faster version of LSTMs, as they use only two gates viz update and output.

However the problem is only partially solved. There still is a sequential path from older cells to the current one. LSTMs even though can learn long term information, but they have trouble remembering sequences of a thousand or more.

\subsection{Transformer}
In the 2017  paper titled ‘Attention is all you need’ by Vaswani et al \cite{vaswani}, they introduce the transformer, a neural network architecture based on self-attention that performed exceptionally well in various language understanding tasks. Unlike RNNs, Transformers do not require the sequential data to be processed in order. Due to this feature the advantage of the Transformer was threefold it allowed for parallelization of tasks, resulted in simpler operations and increased the overall performance of the model on various tasks.
The transformer model is completely reliant on the attention mechanism which allows modeling of dependencies without concern of their distance in the input or output sequences. Transformers forgo the use of recurrence by entirely relying on self-attention thus enabling parallelization.

The basic building blocks of a transformer consists of an encoder and decoder stacked with N(N=6) identical layers. The encoder maps an input sequence (x1,...,xn) to a sequence of continuous hidden context z = (z1,..,zn). The decoder then takes z and generates an output sequence y(y1,..,ym) one element at a time. \\

\textbf{Self Attention:}
An attention function maps a query and a set of key-value pairs all encoded as embeddings to an output which is the weighted sum of the values, where the weight is calculated during training by a compatibility function of the query with the corresponding key. Here keys, values and queries are linearly projected to perform the attention function called the ‘scaled dot-product attention’. The matrix of output is calculated as: 
\[\text { Attention }(Q, K, V)=\operatorname{softmax}\left(\frac{Q K^{T}}{\sqrt{d_{k}}}\right) V\]

\textbf{Multi Headed Attention:}
In the original implementation Vaswani et al found it more beneficial to use self attention h times with different learned linear projections to dk, dq and dv dimensions. This function is performed in parallel yielding a dv dimensional vector. The multi-head characteristic makes it possible to use  different representation subspaces at different positions. 
The multi-headed attention is applied into the transformer in three different ways:
\begin{enumerate}
    \item The encoder makes use of self attention in which the encoder attends to all positions in the previous layer of the encoder.
    \item In a similar way self attention layers in the decoder allow each position in the decoder to attend to all positions in the decoder up to and including that position. 
    \item In encoder-decoder attention as the queries come from the previous decoder layer and the key and values come from output of the encoder, this allows every position in the decoder to attend over all positions in the input sentence. \\
\end{enumerate}

\textbf{Positional Encodings:}
Since the model does not make use of any recurrence layers or convolutions, it needs to have a sense of the sequence order. In order to give the model an understanding of the absolute position of tokens in the sequence, the input embeddings are injected with positional embeddings with which they are summed up. They use two functions to be calculated for each position - sine and cosine. Using these the authors create a unique vector for each position in the sequence.   \\

\textbf{Fully Connected Feed Forward Networks:}
Each layer of the encoder and decoder consists of a Fully Connected Feed Forward network in addition to the attention layer. 

\section{Models}
\subsection{BERT}

BERTs architecture is a multi layer bidirectional Transformer encoder which has the base Transformer by Vaswani et al  as its discrete building block. The original paper by Devlin et al \cite{devlin2018bert} proposes two models viz $BERT_Base$ and $BERT_Large$ with variations with respect to number of layers, hidden size and number of self attention heads. The $BERT_Base$ is a smaller version similar to the size of the OpenAI GPT with 12 number of layers, 768 Hidden states and a total of 12 attention heads while $BERT_Large$ comprises of 24 layers, 1024 Hidden states and 16 attention heads. 

\begin{figure*}
    \centering
    \includegraphics[width=15cm]{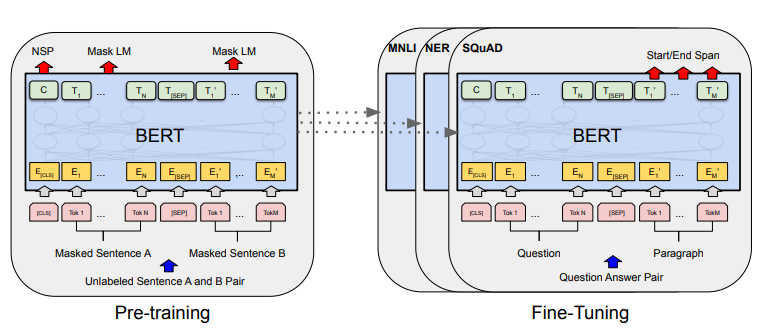}
    \caption{BERT Architecture}\cite{devlin2018bert}
    \label{}
\end{figure*}

\textbf{Pretraining Bert: }
Unlike traditional language modelling algorithms which are trained either from left-to-right or right to left, BERT is trained using two unsupervised tasks i.e Masked Language Modeling and Next Sentence Prediction respectively. These tasks are performed on the BookCorpus (800 million words) and English Wikipedia (2500 million words) datasets.

\textbf{Masked Language Modeling (MLM) :} In order to capture deep bidirectional representation, devlin et al let BERT predict masked words based on the context. They arbitrarily conceal 15\% of the tokens in each sequence. Of these 15\%, only 80\% are substituted with a [MASK] token, 10\% are replaced with a random other token and 10\% are kept the same. In this case the final hidden vectors analogous to the mask tokens are def into an output softmax over the vocabulary.

\textbf{Next Sentence Prediction (NSP) :} In order to train the network to model the relationship between sentences, Devlin et al decided to apply a Next Sentence Prediction task. BERT has to resolve for pairs of sentence segments(each segment can consist of numerous sentences) whether they actually succeed each other or not. Specifically while training, when choosing two sentences, the second sentence follows the first 50\% of the times (labelled as IsNext) and rest of the times is replaced with a random sentence for the corpus(labelled as NotNext)
\subsection{RoBERTa}
In July 2019, Liu et al \cite{liu2019roberta} published a model called RoBERTa (which stands for Robustly-optimized BERT approach). The authors saw room for improvement in the pre-training, arguing that BERT was significantly undertrained. They changed some pre-training configurations, and RoBERTa outperformed BERT on the GLUE benchmark. Since RoBERTa was developed based on BERT, the two models share the transformer architecture. However, Liu et al. examined some of BERT’s pre-training settings in an experimental setup and subsequently decided to implement the following changes:
\begin{enumerate}
    \item \textbf{Adding extra Training Data:} As BERT model was trained on the BookCorpus and English Wikipedia with the total size of training being 16GB, for RoBERTa Liu et al used those two datasets along with three additional sources i.e CC-News, Stories and the OpenWebText dataset making the training data tenfold of the original training data.
    \item \textbf{Dynamic Masking Pattern: }BERT’s masking approach relied on performing masking once during data pre-processing, which resulted in one static mask. Liu et al. adjusted this method when trying different versions of masking for RoBERTa: first, they duplicated the data 10 times over 40 epochs of training. This resulted in four masks, where each training sequence was seen only four times during training procedure (once per mask) instead of every time within one mask. Then, the authors also compared this approach with dynamic masking, where the mask is generated every time the sequence is passed to the model.
    \item \textbf{Removing Next Sentence Prediction Objective :  }Removing the NSP loss led to improvements while testing on four different tasks. Doc-Sentences performed best, but resulted in an increased batch size and less comparability - therefore, Liu et al. decided to use Full-Sentences in the remainder of their experiments.
    \item \textbf{Training with larger batches : }  The original BERT (BASE) was trained over 1 million steps with a batch size of 256 sequences. Liu et al. note that training over fewer steps with an increased batch size would be equivalent in terms of computational costs. Experimenting with those settings, they find out that training with bigger batches over fewer steps can indeed improve perplexity for the masked language modeling pattern and also the final accuracy.

\end{enumerate}
\subsection{DistilBERT}

Sanh et al \cite{sanh2019distilbert} introduce the DistilBERT model which leverages knowledge distillation to pretrain a smaller general purpose language model which can be fine tuned to deliver performances like its larger counterparts. With DistilBERT sanh et al displayed that it is possible to reach and achieve 97\% of BERTs language understanding capabilities while reducing the size of the BERT model by 40\% while being 60\% faster.

Knowledge distillation is a compression technique in which a smaller model i.e the student model is trained to mimic the behavior of a larger model. A standard training objective involves reducing the cross entropy loss between the models predicted distribution and the one-hot factual distribution of the training labels. 

Sanh et al propose three training losses for DistilBERT viz masked language modeling loss (from the standard MLM training task), cosine embedding loss (to align the directions of the student and teacher hidden vector states) and distillation loss. Triple losses make sure that the student model gets trained properly. 
DistilBERT in essence has the same architecture as BERT. The token-type embeddings and pooler are removed while the number of layers is reduced by a factor of 2. DistilBERT follows the footsteps of Roberta and avoids Next Sentence Prediction task and is trained on a very large batch size (upto 4K examples per batch) using dynamic masking. 

\subsection{Albert}
While RoBERTa focused on enhancing performance of BERT and DistilBERT focused on efficiency and speed, ALBERT (A Lite BERT) is built to address both. ALBERT achieves SOTA performance with lower memory usage and increased training speed in comparison to BERT. Since BERT is parameter inefficient Lan et al \cite{lan2019albert} in their paper address this issue by applying techniques to reduce parameters to 1/10th of the original model without substantial performance loss. The authors experiment with two methods to reduce the model size: 1. Factorized embedding parametrization and cross-layer parameter sharing. In addition, they improve the model training by modifying the NSP objective with sentence order prediction. 

\textbf{Factorized Embedding Parameterization: } 
In BERT and BERT based models, the size of the BertWordPiece embeddings and the size of the hidden layers is linked. Lan et al argue that this is inefficient, because the two have different purposes. WordPiece Embeddings supposedly capture the general meaning of the words which is context independent, whereas hidden layers map the meaning dependent on the specific context. Hidden layers therefore are much more intricate: they need to store more information and be updated more often during the training process. In ALBERT, Lan et al. decompose the large vocabulary matrix into two smaller matrices. Thereby, they separate the size of the layers and reduce the number of parameters.

\textbf{Cross-Layer Parameter Sharing: } 
 Another way to reduce parameters is to enable sharing of parameters across layers. Lan et al try three different settings for parameter sharing: sharing the feed forward network parameters, the attention parameters and sharing both at the same time. Parameter sharing prevents the number of parameters to grow with the depth of a network.

\section{Results}

\begin{table}[h]
\begin{tabular}{llllll}
\hline\\
\textbf{Metri}c & \textbf{Accuracy} & \textbf{Precision} & \textbf{Recall} & \textbf{F1 Score} & \begin{tabular}[c]{@{}l@{}}\textbf{AUC}\end{tabular} \\ \hline \\
\textbf{BERT}         & 0.9691   & 0.7629    & 0.7271 & 0.7446   & 0.9796                                                           \\ \\\hline\\
\textbf{DISTILBERT}   & 0.9663   & 0.7481    & 0.6854 & 0.7154   & 0.9758                                                           \\ \\\hline\\
\textbf{ROBERTA}      & 0.9551   & 0.6869    & 0.5030 & 0.5807   & 0.9484                                                           \\ \\\hline\\
\textbf{ALBERT}       & 0.9495   & 0.6583    & 0.3827 & 0.4840   & 0.9239                                                           \\ \\\hline\\
\end{tabular}
\end{table}

Because of a severe imbalance between the class labels with the questions pertaining to sincere labels being of a higher percentage than Insincere questions, traditional metrics of validating the model do not capture the essence of the model in its entirety. Hence we use metrics such as the AUC score along with F1 score to capture the holistic efficiency of the trained models.

\textbf{AUC score :} AUC score is a performance measure of classification models at various threshold levels. It tells how much the model is capable of distinguishing classes. It plots the false positive rate against the true positive rate of the model. An ideal model would have an AUC score of 1.  \\

\textbf{F1 Score :} F1 score of a classification model is defined as the harmonic mean of the precision and recall. Precision is the number of correctly identified true labels divided by all true labels in the data. Recall is the number of correctly identified true labels divided by number of true samples that should have been identified as positive. Higher the F1 score better the performance of the model.

\section{Conclusion}

In this paper, we aimed to identify Insincere Questions from text state of the art NLP models. Starting off with simple methods to the most cutting edge, we illustrated how NLP models can compete with others. In order to do so, we explored how BERT and three BERT-based transformer models approach text classification. RoBERTa, DistilBERT, and ALBERT each improve the original model in a different way with regards to performance and speed. In our application, we demonstrated the easiest way to implement transformer models, how to modify the standard settings and what else to pay attention to. On the task of identifying insincere user intent BERT performed best. However, the field of NLP is fast moving - and we are excited to see what the next transformational generation of models will bring.

\bibliographystyle{unsrtnat}
\bibliography{bibliography}

\end{document}